\documentclass[10pt,twocolumn,letterpaper]{article}

\usepackage{iccv}
\usepackage{times}
\usepackage{epsfig}
\usepackage{graphicx}
\usepackage{amsmath}
\usepackage{amssymb}
\usepackage{subcaption}
\usepackage{bbm}

\iccvfinalcopy

\usepackage[breaklinks=true,bookmarks=false]{hyperref}

\def\iccvPaperID{74} 

\ificcvfinal\fi

\begin{document}

\title{Self-Supervised Multiple Instance Learning \\for Acute Myeloid Leukemia Classification}


\author{Salome Kazeminia*\\
TUM\\
Helmholtz Munich\\
{\tt\small salom.kazeminia@helmholtz-muenchen.de}
\and
Max Joosten*\\
Eindhoven University of Technology\\
Helmholtz Munich\\
{\tt\small m.p.h.joosten@student.tue.nl}
\and
Dragan Bosnacki\\
Eindhoven University of Technology\\
{\tt\small d.bosnacki@tue.nl}
\and
Carsten Marr\\
Helmholtz Munich\\
{\tt\small carsten.marr@helmholtz-muenchen.de}
}

\maketitle
\ificcvfinal\thispagestyle{empty}\fi

\begin{abstract}
Automated disease diagnosis using medical image analysis relies on deep learning, often requiring large labeled datasets for supervised model training. 
Diseases like Acute Myeloid Leukemia (AML) pose challenges due to scarce and costly annotations on a single-cell level. 
Multiple Instance Learning (MIL) addresses weakly labeled scenarios but necessitates powerful encoders typically trained with labeled data. In this study, we explore Self-Supervised Learning (SSL) as a pre-training approach for MIL-based AML subtype classification from blood smears, removing the need for labeled data during encoder training. 
We investigate the three state-of-the-art SSL methods SimCLR, SwAV, and DINO, and compare their performance against supervised pre-training. 
Our findings show that SSL-pretrained encoders achieve comparable performance, showcasing the potential of SSL in MIL. 
This breakthrough offers a cost-effective and data-efficient solution, propelling the field of AI-based disease diagnosis.
\end{abstract}

\section{Introduction}
The precise classification of Acute Myeloid Leukemia (AML) genetic subtypes is imperative for selecting appropriate treatment modalities and improving patient outcomes. 
Achieving such accurate classification entails the utilization of genetic testing to discern distinct AML genetic subtypes~\cite{Papaemmanuil2016GenomicLeukemia}, albeit with inherent cost and complexity.
In this context, blood smear microscopy emerges as a compelling alternative method, entailing a scrupulous microscopic examination of stained blood samples to reveal elusive malignant cells~\cite{yu_leukemia}. 
However, the intrinsic rarity of these malignant cells presents a formidable challenge for clinicians undertaking this diagnostic endeavor.

Automated disease diagnosis through image-based approaches has garnered remarkable attention in research and clinical spheres~\cite{matek2019human,matek2021highly,sidhom2021deep,eckardt2022deep,eckardt2022deep}. 
However, in the context of AML diagnosis, a particular challenge arises due to the availability of weakly labeled data.
In this scenario, each diagnosis relies on patient-specific leukocyte images, forming part of an extensive image set that also encompass non-relevant cells. 
This weak data labeling poses an intricate task for achieving accurate classification, thereby necessitating innovative methodologies to surmount the constraints imposed by the limited availability of precisely labeled information.
To surmount this hurdle, Multiple Instance Learning (MIL) emerges as a compelling, weakly supervised method. 
Hehr et al.\ elucidates the remarkable performance of MIL by emphasizing class-specific leukocyte images detected through an attention mechanism~\cite{hehr2023explainable}. 
The triumph of MIL, however, hinges significantly upon the caliber and efficacy of the representations acquired by its encoder model. 
When confronted with insufficient training data to effectively train the encoder, pre-training emerges as a viable alternative~\cite{hehr2023explainable}. 
However, accessing fully-supervised datasets can be challenging, especially in medical environments where data privacy and scarcity concerns often arise. 

\begin{figure*}[t!]
	\begin{center}
		\includegraphics[width=\textwidth]{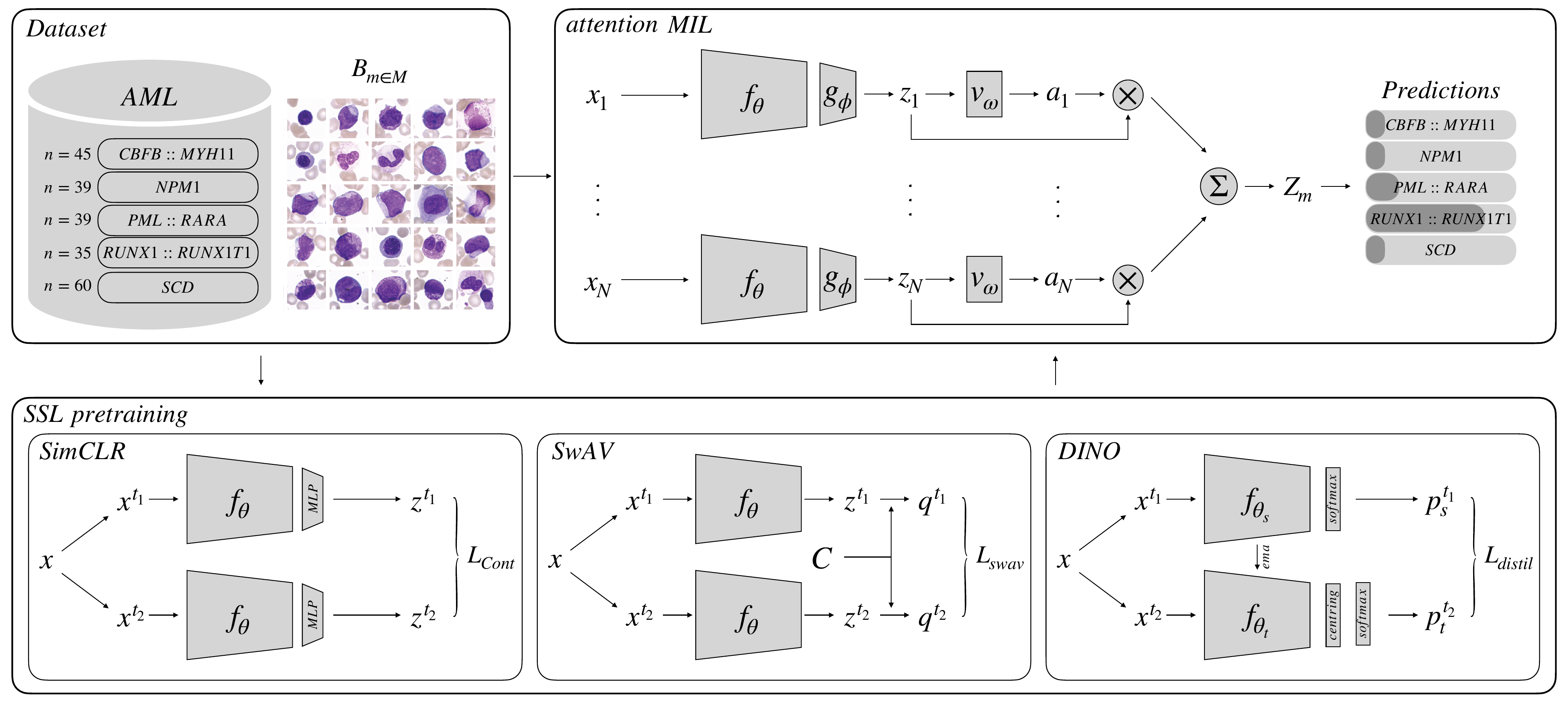}
	\end{center}
	\caption{Overview of our data-efficient MIL model for AML genetic subtypes classification. We pre-train an MIL encoder $f_{\theta}$ with one of the state-of-the-art SSL models, i.e., SimCLR, SwAV, and DINO. Then we embed the trained encoder in the attention MIL architecture. For both SSL and MIL the same training dataset is used.}
	\label{fig:diagram}
\end{figure*}

Self-Supervised Learning (SSL) has proved to be a promising pre-training method for Multiple Instance Learning (MIL) classifiers, particularly in data-limited scenarios like medical environments where obtaining extensive labeled datasets is challenging~\cite{li2021dual,Lu_2023_CVPR}. 
Unlike supervised pre-training methods that rely on costly annotations, SSL techniques enable the encoder to learn informative representations from abundant unlabeled data without explicit annotations. 
By leveraging the inherent structure and relationships within the data, SSL empowers the encoder to capture useful domain-specific features, enhancing the performance and generalizability of MIL models.

In light of annotated data limitation, we turned to Self-Supervised Learning (SSL) techniques as a compelling solution, enabling the encoder to learn meaningful representations from vast amounts of unlabeled cells in without needing explicit annotations. 
This strategic shift towards SSL not only mitigates the reliance on fully-labeled datasets but also empowers our model to capture more robust and domain-adaptive features of MIL frameworks~\cite{krishnan2022self,ciga2022self,chen2022scaling}. 

In this work we investigate three SSLmethods that utilize distinct strategies to address classification of AML genetic subtypes within the framework of MIL. 
Specifically, we conducted experiments using SimCLR, SwAV, and DINO, which represent state-of-the-art performance in contrastive-instance, clustering-based, and self-distillation-based SSL approaches, respectively, and have demonstrated promising performance on widely recognized computer vision benchmark datasets, such as ImageNet~\cite{russakovsky2015imagenet}. 
By benchmarking these SSL techniques, we aim to assess their performance and suitability for AML subtype classification in the context of MIL (see Figure \ref{fig:diagram}). 
This validation highlights the efficacy and promise of leveraging SSL techniques for accurate and data-efficient AML subtype classification within the MIL framework without tedious single-cell annotation tasks.

\section{Related Work}

\subsection{Self-supervised learning models}
The primary objective of the SSL method is to acquire meaningful representations from unlabeled data, thereby circumventing the need for explicit annotations. 
Several SSL models have been introduced, each characterized by ingenious methodologies aimed at harnessing the wealth of information present in large-scale, unlabeled datasets.
However, mode collapse is a critical challenge in SSL methods, where learned features collapse into a single point or limited space, hindering model generalization and downstream task performance~\cite{jing2021understanding}. 
Addressing mode collapse is crucial to ensure the effectiveness and robustness of SSL approaches.

{\bf Simple framework for contrastive learning of visual representations (SimCLR)}~\cite{Chen2020ARepresentations} employs a contrastive learning framework to learn meaningful and transferable image representations. 
For each image $x_{i} \in X$, diverse augmentations $t \in T$ are applied, resulting in augmented views $x_{i}^{t_1}$ and $x_{i}^{t_2}$ and an encoder function followed by a projection head maps them to latent representations ${z_{i}^{t_{1}}}, {z_{i}^{t_{2}}}$.
The contrastive loss function used in SimCLR aims to maximize the cosine similarity ($Sim$) between the projection feature vectors of positive pairs (augmented views of the same image), while minimizing the similarity between the projection feature vectors of negative pairs (augmented views of different images, indicated with $\mathbbm{1}_{[k\neq i]}$, which equals $1$ if $k\neq i$ and $0$ otherwise).
Mathematically, the contrastive loss function can be defined as follows:
\begin{equation}
	\label{eq:simclr_loss}
	L_{Cont}=-\log{\frac{\exp(Sim({z_{i}^{t_{1}}},{z_{i}^{t_{2}}})/\tau)}{\sum_{k=1}^{2N} {\mathbbm{1}_{[k\neq i]}}\exp(Sim({z_{i}^{t_{1}}},{z_{k}^{(t)}})/\tau)}}.
\end{equation} 
In this equation, $N$ represents the number of instances in the mini-batch, ${z_{i}^{t_{1}}}, {z_{i}^{t_{2}}}$ are the projection feature vectors of the positive pairs (augmented views of the same image), $Sim({z_{i}^{t_{1}}}, {z_{i}^{t_{2}}})$ denotes the cosine similarity between these vectors and$\tau$ is the temperature parameter, which scales the similarity values.
The contrastive loss encourages the projection feature vectors of positive pairs to have high cosine similarity, making them closer in the embedding space, while the similarity between the projection feature vectors of negative pairs is minimized, pushing them further apart.
This way, the model learns to create compact clusters for instances belonging to the same image (positive pairs) and maximize the separation between instances from different images (negative pairs), leading to more meaningful and discriminative feature representations.

{\bf Swapping Assignments between Views (SwAV)}~\cite{caron2020unsupervised}  is a clustering-based SSL approach, where visual features are learned by contrasting the cluster assignments of different augmentations $t \in T$ applied to the same image.  
The encoder network maps these augmented views into encoded feature vectors $z^{t_1}$ and $z^{t_2}$, respectively. 
Prototype vectors representing cluster centroids are maintained as $c_j$ for each cluster $j \in J$, where $J$ is the set of all clusters. 
The objective loss function is formulated as follows
\begin{equation}
	\label{eq:swav_loss}
	L_{swav}(z^{t_1},z^{t_2}) = l(z^{t_{1}},q^{t_{2}}) + l(z^{t_{2}},q^{t_{1}})
\end{equation}
where $q^{t_{1}}$ and $q^{t_{2}}$ are abstract information of latent representations $z^{t_{1}}$ and $z^{t_{2}}$ correspondingly. 
The loss function $l(.)$ measures the fit between representations of two augmentation of image as
\begin{equation}
	l(z^{t_{1}},q^{t_{2}}) = -\sum_{i \in J}^{}q^{t_{2}}_{(i)}\log{p_{(i)}^{t_{1}}},
\end{equation} 
where 
\begin{equation}
	p_{(i)}^{t_{1}} = \frac{\exp{(z^{t \in T}c_{i}/\tau)}}{\sum_{j \in J}^{}\exp{(z^{t \in T}c_{j}/\tau)}}.
\end{equation}
To obtain cluster assignments for the encoded feature vectors, a softmax function with a temperature parameter $\tau$ is applied, resulting in cluster assignment probabilities $p_{(k)}^{t_{1}}$ and $p_{(k)}^{t_{2}}$. 
SwAV then swaps these assignments to form new assignment pairs, represented as $(p_{ij}^{t_1}, p_{ik}^{t_{2}})$ and $(p_{ij}^{t_{1}}, p_{ik}^{t_1})$, where $k \in J$ represents a different cluster. 
The prototype vectors $c_{j}$ are updated using these swapped assignment pairs, aiming to maintain balanced assignments and ensure each feature vector is associated with a single prototype with the highest probability. 
This fosters a diverse and well-balanced representation of clusters, mitigating the risk of collapsing into a few dominant clusters. SwAV's approach enables the learning of robust visual features without the need for explicit data labeling, improving model generalization performance.

{\bf Self-distillation with no labels (DINO)}~\cite{caron2021emerging} leverages the concept of self-distillation, where a teacher network $f_{\theta_{t}}$ guides the learning of a student network $f_{\theta_{s}}$ in a self-supervised manner. 
The main idea is to train the student network to mimic the output of the teacher network, which helps the student network learn useful representations without the need for explicit labels.
During training, different augmentations of an image $x \in X$ pass through student and teacher networks and a distillation loss encourages the student to approximate the teacher's predictions as
\begin{equation}
	\mathcal{L}_{\text{distill}} = -\sum_{i=1}^{K} p_{t}^{i} \log(p_{s}^{i}),
\end{equation}
where $p^{i}(x)$ represents the probability of image $x$ falling into soft-class $i$. These probabilities are calculated as:
\begin{equation}
	p_{s\vee t}^{i}(x)=\frac{\exp{(f^{i}(x^{t};\theta_{s\vee t}/\tau})}{\sum_{k=1}^{K}\exp(f^{k}(x^{t};\theta_{{s\vee t}})/\tau)},
\end{equation} 
where $K$ is the number of soft-classes referring to a continuous and smooth representation of the distribution of each data and its augmentations.
Backpropagation is applied only to the student network, while the teacher network is updated using the exponential moving average of the student's parameters, allowing knowledge transfer from student to teacher.
To prevent mode collapse and improve the learning process, DINO introduces two additional techniques: centering and sharpening. Centering is applied to the teacher network and involves subtracting the average of projection vectors from individual vectors to balance predictions. 
Sharpening uses a lower temperature ($\tau$) in the softmax layer for the teacher network, making the prediction task more challenging for the student network.
	
\section{Overview of the approach}
Our framework, similar to the approach proposed in \cite{hehr2023explainable}, is specifically designed for analyzing microscopic images within a Multiple Instance Learning (MIL) setting. 

In this context, bags denoted as ${B_{1}, ..., B_{M}}$ represent sets of blood sample images containing white blood cells, where each individual cell is represented as ${x_{1}, ..., x_{N}} \in B_{m\in M}$. 
For each bag, we have one expert-annotated label denoted as $y_m$, indicating the specific subtype of Acute Myeloid Leukemia (AML) associated with the corresponding blood sample.

Notably, the framework is customized to meet the demands of this particular domain, where cells are distributed along independent spatial locations within the image.
This is different from cases where tiling techniques are employed to capture cells~\cite{shao2021transmil}.
\subsection{Feature extraction}

An encoder is utilized to extract class-related features from individual cells, which are then mapped to $k$-dimensional feature vectors denoted as ${z^{k}_{1}, ..., z^{k}_{N}}$. 
The encoder comprises a pre-trained ResNet 18~\cite{he2016deep} with frozen weights. 
For pre-training, the encoder is incorporated into a Self-Supervised Learning (SSL) architecture, where the training data is passed through the network without any labels, enabling the encoder to learn to map similar cells and their augmentations close to each other in the feature space. 
Additionally, a four-convolution layer network $g(.)$ with learnable parameters $\phi$ is introduced and trained with the Multiple Instance Learning (MIL) loss. 
This network takes the feature vectors $z^{k}_{n}$ and outputs $z^{k^{\prime}}_{n}$ in a lower-dimensional space $k^{\prime}$ with $k^{\prime} < k$. 
The combination of the encoder and the convolutional network allows for effective feature extraction and dimension reduction, facilitating accurate and efficient classification within the MIL framework.
\subsection{Attention pooling}
A multi-head deep attention network, denoted as $v_{\omega}$, is equipped with two linear layers to estimate the attention scores $a^{c}_{n}$, for each cell's feature vector $z^{k^{\prime}}_{n}$ with respect to each class $c \in C$, where $C$ denotes the set of all classes. 
The attention scores are computed as 
\begin{equation}
	\label{attention}
	a^{c}_{n} = v_{\omega}(z^{k^{\prime}}_{n}), 
\end{equation}
Subsequently, the final representation of the bag $B_{m}$ is calculated as follows:
\begin{equation}
	Z^{c}_{m} = \sum_{n=1}^{N} a_{n}^{c} \cdot z^{k^{\prime}}_{n},
\end{equation}
where $Z^{c}_{m}$ denotes the representation of bag $B_{m}$ for class $c$, obtained by combining the attention scores $a_{n}^{c}$ with the lower-dimensional feature vectors $z^{k^{\prime}}_{n}$.

This representation $Z^{c}_{m}$ is then utilized for the final multi-class prediction, and the objective function $L_{\text{MIL}}$ is defined as:
\begin{equation}
	L_{MIL} = CE(f_{MIL}(Z^{k^{\prime}}_{m};\phi, \omega), y_{m}),
\end{equation}
where $CE$ represents the categorical cross-entropy loss and $f(.)$ denotes the classifier head function. 
The overall MIL framework leverages this mechanism to predict the AML subtype for each blood sample, utilizing both the attention scores and the lower-dimensional feature vectors to make accurate and explainable predictions.
This formulation enables the identification of the most contributing cells based on the class-specific attention scores, allowing for cell-level explanation and providing valuable insights into the model's decision-making process, which is particularly crucial for medical applications.

\section{Experiments}

\begin{figure*}[!t]
	\begin{center}
		\includegraphics[width=\textwidth]{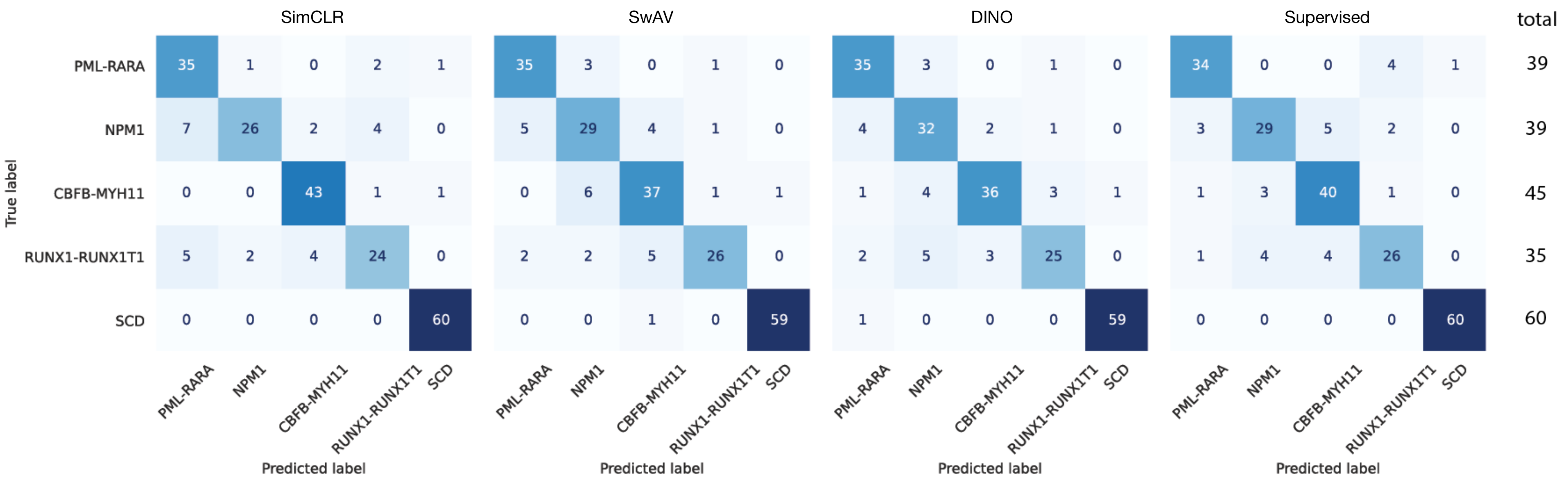}
	\end{center}
	\caption{The confusion matrix presents the test fold results of the first run for each pre-training method. While different SSL pre-training methods lead to varying class-wise performance, overall SSL pre-training performs correspondingly to fully supervised pre-training}
	\label{fig:CM}
\end{figure*}

\subsection{Dataset}
The AML dataset utilized in this study is publicly available, introduced by Hehr et al.\ in their work on explainable AI for AML genetic subtype classification \cite{hehr2023explainable} (\href{https://wiki.cancerimagingarchive.net/pages/viewpage.action?pageId=145753815}{AML dataset}). 
The dataset comprises $242$ blood smears from the Munich Leukemia Laboratory (MLL), which were associated with four distinct AML genetic subtypes: APL with "$PML::RARA$ fusion" $(n = 45)$, AML with "$NPM1$" mutation $(n = 39)$, AML with "$CBFB::MYH11$" fusion (without NPM1 mutation) $(n = 39)$, and AML with "$RUNX1::RUNX1T1$ fusion" $(n = 35)$. 
Additionally, healthy stem cell donors (SCD) were included as controls $(n = 60)$. 
The ground truth labels of AML genetic subtypes was determined through genetic testing.

Each blood smear was stained and scanned, resulting in a total of $218$ patients and $101,947$ single-cell images with a resolution of $144 \times144$ per blood smear, corresponding to a size of $24.9\mu m \times 24.9\mu m$. 
Furthermore, $1,983$ single-cell images were expertly annotated by a cytologist, allowing for experiments using single-cell labels.
This comprehensive dataset facilitated the evaluation of our MIL-based classification framework for accurate and interpretable AML subtype classification.

\subsection{Training}
In our experiments, we conduct $5$-fold stratified cross-validation for both the SSL encoder pre-training and the MIL head training. 
The dataset is divided into $60\%$ for training, $20\%$ for validation, and $20\%$ for testing. 
To account for the uncertainty introduced by the MIL training procedure, we repeated the training of the MIL head three times with different random seeds.

For the supervised encoder, we adopt a ResNet $34$ model pre-trained on an external dataset with $300,000$ annotated single-cell images comprising $23$ different classes. 
The pre-training process in \cite{hehr2023explainable} involved random rotation, rescaling, translation, random erasing, and horizontal/vertical flipping, while oversampling was utilized to handle class imbalance. 
The model was trained for $50$ epochs, and the stochastic gradient descent (SGD) optimizer with a learning rate of $5\times 10^{-4}$ was employed.

The three SSL methods, SimCLR, SwAV, and DINO, are fine-tuned with specific hyperparameters for encoder pretraining. 
For {\bf SimCLR}, we use SGD with momentum, learning rate scaling, and cosine annealing during training, along with Large Batch Training of Convolutional Networks (LARC)~\cite{you2017large} to improve model convergence. 
The contrastive loss function is employed with a temperature of $0.1$, and a two-layer projection head with dimensions $[512, 512]$ and $[512, 64]$. 
Random augmentations such as rotation, resize cropping, flipping, color jittering, greyscale, and Gaussian blur are applied to the input images.

Similarly, for {\bf SwAV}, we utilize SGD with momentum, learning rate scaling, and cosine annealing, trained for $500$ epochs with a batch size of $512$, and LARC was used for convergence enhancement. 
The projection head comprised a $3$-layer MLP with specific dimensions and $300$ prototype vectors were updated with the Sinkhorn-Knopp algorithm~\cite{cuturi2013sinkhorn}.
Augmentations included multi-crops with global and local views, vertical/horizontal flipping, Colorjitter, and Gaussian blurring.

For {\bf DINO}, we used the AdamW~\cite{loshchilov2017decoupled} optimizer with momentum and a weight decay scheduler, batch size of $512$, and LARC for convergence. 
Learning rate scaling and cosine annealing is applied, and the teacher model is warmed up with a teacher temperature of $0.04$, increasing to $0.07$.
The momentum of the momentum encoder is adjusted during training. 
Similar augmentations as SwAV is applied, with $2$ global and $8$ local crops generated from a single image.

The {\bf MIL} model is fine-tuned for $50$ epochs or until the validation loss plateau for $20$ epochs. 
We utilize the SGD optimizer with Nesterov momentum~\cite{nesterov2003introductory} and set the learning rate to $0.015$. 
To ensure stable training in later epochs, the learning rate is gradually reduced using cosine annealing. 
Back-propagation is performed only after gradients are accumulated for $10$ batches. 
The batch size is determined based on the number of instances associated with each patient, with a maximum of $500$ instances per batch.
During training, we apply random vertical and horizontal flipping as augmentations to the instances in each bag. 
To address the class imbalance between the AML subtypes, we employ a higher sampling rate for bags containing the minority class. 
This approach results  in under-sampling of majority class bags and over-sampling of minority class bags, effectively balancing the class distribution and enhancing the model's performance in handling imbalanced data.

We implement both the SSL pre-training and the classifier in PyTorch, utilizing the VISSL self-supervised learning framework~\cite{goyal2021vissl} for pre-training. 
All experiments are conducted on a single desktop-class NVIDIA RTX $3090$ GPU. 
To ensure robustness and reliability, we run each pre-trained model three times, and with 5-fold cross-validation, resulting in $15$ sets of predictions. 
The performance metrics are reported as the mean and standard deviation of these prediction sets.

To reduce CPU overhead during training, the augmentation pipelines are implemented using albumentations~\cite{buslaev2020albumentations}. 
Additionally, a combination of mixed precision training and activation checkpointing techniques is employed to accelerate the training process and reduce the memory footprint, ensuring efficient utilization of computational resources.

\begin{figure*}[t!]
	\begin{center}
		\includegraphics[width=\textwidth]{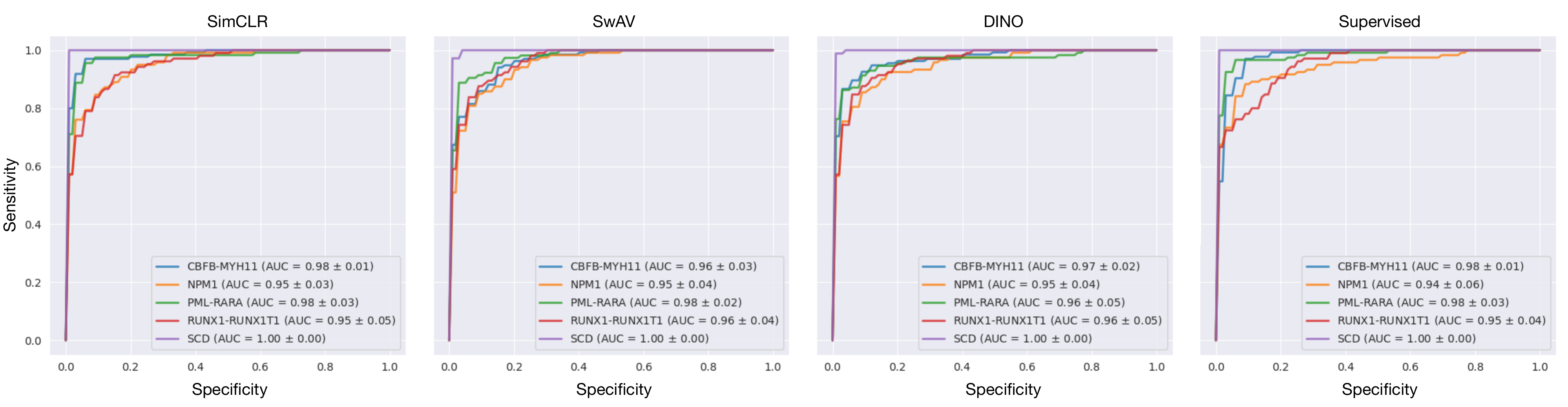}
	\end{center}
	\caption{Similar mean ROC curves for MIL model resulting from different pre-training methods. Differences in the AUC score for genetic subtype prediction is within the margin of error. }
	\label{fig:AUC}
\end{figure*}

\begin{figure*}[t!]
	\begin{center}
		\includegraphics[width=\textwidth]{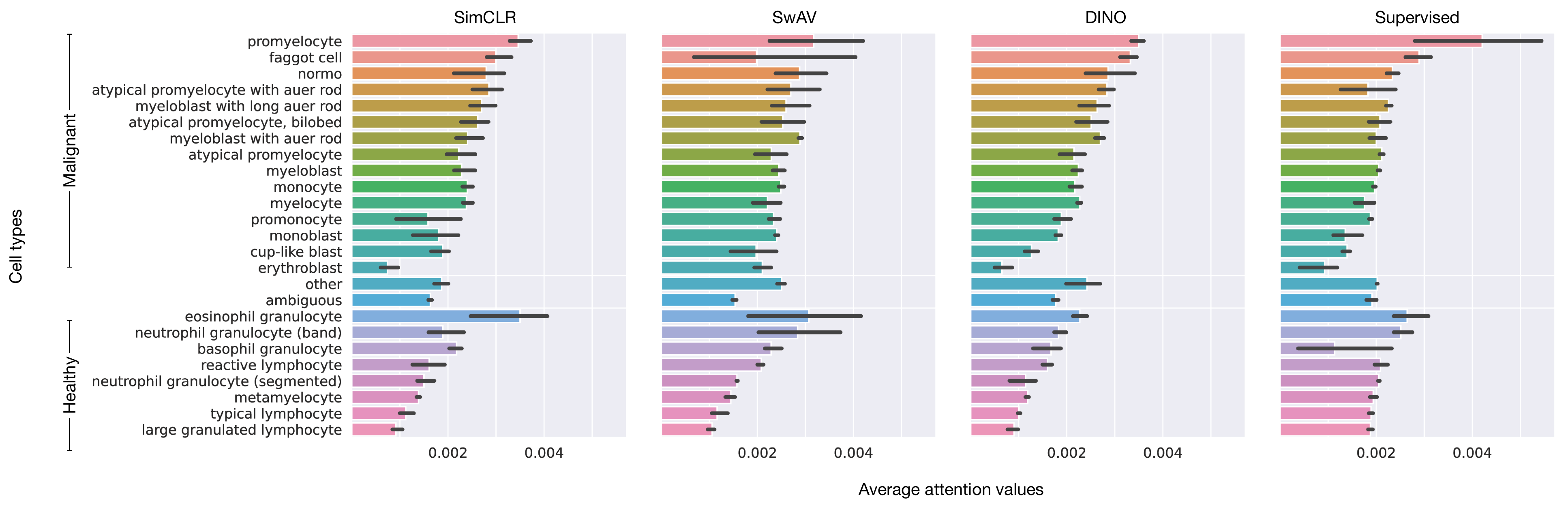}
	\end{center}
	\caption{Averaged attention values for all labeled instances per model. Most of cell types with higher attention are malignant, while most lower attention cell types a healthy. SimCLR and DINO pretrained models provide more robust attention scores for malignant cells compared to the supervised pre-training.}
	\label{fig:att}
\end{figure*}

\subsection{Quantitative results}
The MIL classifier, when trained on an SSL pre-trained encoder, exhibits comparable performance to the classifier trained on a supervised encoder in classifying AML subtypes (see Figure \ref{fig:CM}). 
The macro-averaged classification metrics, as shown in Table \ref{tab:F1}, are very similar across the pre-training methods. 
For instance, the classifier trained on the supervised encoder achieves high classification performance (F1-macro $0.86 \pm 0.09$), and this level of performance is matched by the SimCLR encoder (F1-macro $0.85 \pm 0.04$). 

\begin{quote}
	\begin{center}
		\begin{table}[htp!]
			\centering
			\begin{tabular}{|l|l|l|l|l|l|}
				\hline
				Pre-training & F1 & ROC AUC & PR AUC \\
				\hline\hline
				DINO &  0.85 ± 0.07 & 0.97 ± 0.03 & 0.91 ± 0.08 
				\\
				SimCLR &  0.85 ± 0.04 & \textbf{0.97 ± 0.02} & \textbf{0.92 ± 0.05} 
				\\
				SwAV &  0.82 ± 0.06 & \textbf{0.97 ± 0.02} & 0.91 ± 0.05 
				\\
				Supervised &  \textbf{0.86 ± 0.09} & 0.97 ± 0.03 & 0.92 ± 0.07
				\\
				\hline
			\end{tabular}
			\caption{Averaged classification performance of MIL pre-trained with different SSL methods vs. supervised learning methods. Results are averaged over $3$ runs over $5$-fold cross validation and reported in mean ± s.d. format.}
			\label{tab:F1}
		\end{table}
	\end{center}
\end{quote}

Notably, different pre-training methods exhibit varying classification performances across different classes. 
For instance, SimCLR demonstrates superior performance in predicting $CBFB::MYH11$ fusion, while it performs slightly worse in the classification of $RUNX1::RUNX1T1$ fusion compared to the other pre-training methods.
The average area under the Receiver Operator Characteristic curve (ROC AUC)  and the average area under the Precision-Recall curve (PR AUC) (Figure \ref{fig:AUC}) also align closely with the results from the SimCLR model. 
Detailed classification results for all combinations of models and classes can be found in Appendix A.

\begin{figure*}[t!]
	\begin{center}
		\includegraphics[width=\textwidth]{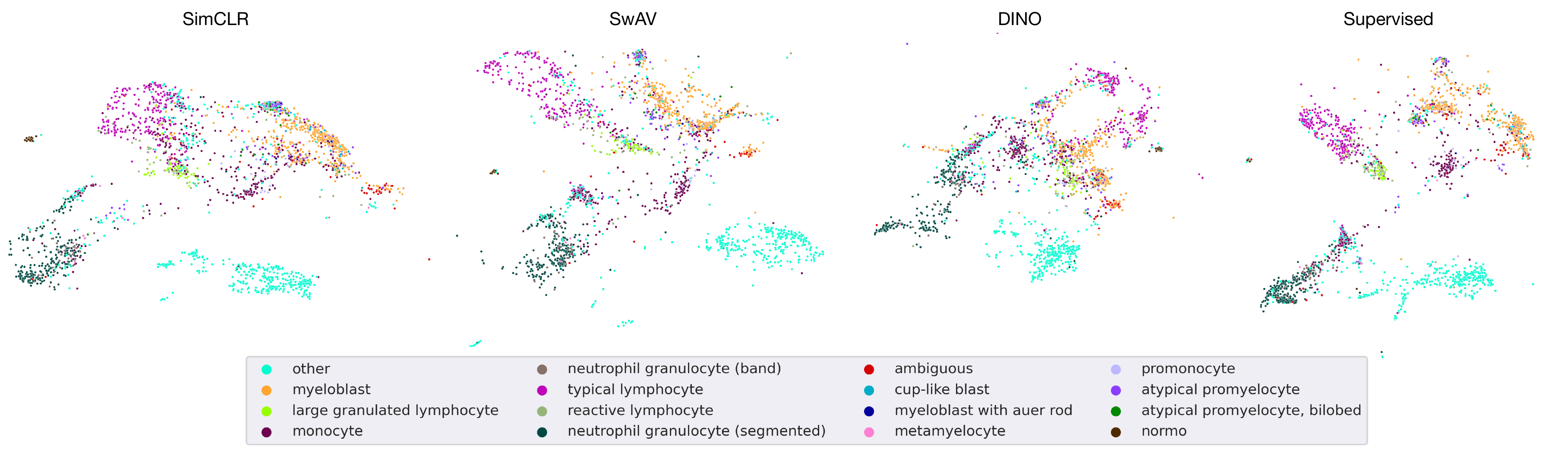}
	\end{center}
	\caption{UMAP embedding visualization of encoder features with expert cytologist-assigned cell labels. SSL pre-trained encoder could distinguish cell-types without any label similar to fully supervised pre-trained encoder.}
	\label{fig:umap}
\end{figure*}

The attention module integrated into the MIL classifier demonstrates its efficacy in focusing on malignant cells. 
These cells consistently receive higher attention values, enhancing the model's interpretability. 
By ranking the input images based on the attention values, the model highlights cells crucial for predictions, even in the absence of per-cell labels. 
Examining the attention patterns of classifiers trained with different pre-trained SSL encoders in Figure \ref{fig:att} reveals that malignant cells consistently obtain higher attention values than healthy cells. 
Notably, the classifier trained with the SimCLR encoder exhibits a larger gap in average attention between malignant and healthy cells compared to the classifier trained with the supervised encoder.
This observation suggests that pre-trained encoders produce features that are more easily ranked by the classifier, resulting in more accurate ranking of malignant cells. 
Consequently, the attention mechanism proves to be a valuable asset in identifying crucial cells for AML subtype classification, contributing to the model's improved performance and interpretability.

\subsection{Qualitative results}
In this work, we delve into the impact of various Self-Supervised Learning (SSL) methods on the encoder's latent space, which plays a crucial role in the Acute Myeloid Leukemia (AML) classification task. 
To assess the quality of the latent space, we employ UMAP~\cite{mcinnes2018umap}, a non-linear dimensionality reduction technique that preserves both local and global data structures. 
The high-dimensional features extracted from the first cross-validation fold for each encoder ($512$ dimensions) are projected into a low-dimensional map ($2$ dimensions) for visualization. We focus on a subset of the data with expert cytologist annotations to gain insights into the latent space's arrangement and structure (see Figure \ref{fig:umap}).
The visualization, based on a subset of expert-annotated cells, demonstrates that all SSL pre-training methods achieve good class separation without labeled data.
Interestingly, the ambiguous cells are clearly separated into their own cluster by SSL pre-training, while the supervised encoder tends to associate them more with the myeloblast cluster. 
This indicates that SSL pre-training effectively generates distinctive features, even for challenging cases, enhancing the encoder's ability to detect such edge cases.

\section{Conclusion}
We explored various self-supervised learning (SSL) methods for encoder pre-training in the context of acute myeloid leukemia (AML) subtype classification. 
We compared three state-of-the-art SSL methods: SimCLR, SwAV, and DINO, to a fully-supervised encoder. 
Our results showed that SSL-based pre-training can yield useful features for AML subclass classification, achieving comparable performance to a fully-supervised approach while using a smaller, unlabeled dataset. 
SimCLR demonstrated the best performance, likely due to its effective contrastive learning framework. 
On the other hand, SwAV's clustering-based approach showed less favorable results, possibly due to the difficulty of distinguishing subtle differences in cytology images. 
Overall, SSL methods have the potential to leverage unlabeled data in medical environments, opening up new possibilities for efficient and effective image classification tasks.

\section{ackknowledgement}
The Helmholtz Association supports the present contribution under the joint research school “Munich School for Data Science - MUDS”.
C.M. has received funding from the European Research Council (ERC) under the European Union’s Horizon 2020 research and innovation program (Grant Agreement No. 866411).

{\small
\bibliographystyle{ieee_fullname}
\bibliography{main}
}
\end{document}